\documentclass[10pt,twocolumn,letterpaper]{article}

\usepackage{cvpr}              

\usepackage{graphicx}
\usepackage{amsmath}
\usepackage{amssymb}
\usepackage{booktabs}
\usepackage{bm}
\usepackage{amsfonts}

\usepackage[pagebackref,breaklinks,colorlinks]{hyperref}

\usepackage[capitalize]{cleveref}
\crefname{section}{Sec.}{Secs.}
\Crefname{section}{Section}{Sections}
\Crefname{table}{Table}{Tables}
\crefname{table}{Tab.}{Tabs.}

\def\cvprPaperID{*****} 
\def\confName{CVPR}
\def\confYear{2023}

\begin{document}

\title{Local Region Perception and Relationship Learning Combined with Feature Fusion for Facial Action Unit Detection}

\author{
Jun Yu,  Renda Li, Zhongpeng Cai, Gongpeng Zhao, Guochen Xie, Jichao Zhu, Wangyuan Zhu \\
University of Science and Technology of China\\
\tt\small harryjun@ustc.edu.cn\\
\tt\small \{rdli,zpcai,zgp0531,xiegc,jichaozhu,zhuwangyuan\}@mail.ustc.edu.cn
}

\maketitle

\begin{abstract}
Human affective behavior analysis plays a vital role in human-computer interaction (HCI) systems. In this paper, we introduce our submission to the CVPR 2023 Competition on Affective Behavior Analysis in-the-wild (ABAW). We propose a single-stage trained AU detection framework. Specifically, in order to effectively extract facial local region features related to AU detection, we use a local region perception module to effectively extract features of different AUs. Meanwhile, we use a graph neural network-based relational learning module to capture the relationship between AUs. In addition, considering the role of the overall feature of the target face on AU detection, we also use the feature fusion module to fuse the feature information extracted by the backbone network and the AU feature information extracted by the relationship learning module. We also adopted some sampling methods, data augmentation techniques and post-processing strategies to further improve the performance of the model.
\end{abstract}

\section{Introduction}
\label{sec:intro}

The affective behavior analysis in-the-wild (ABAW) \cite{kollias2023abaw, kollias2023abaw2} is a major targeted characteristic of human-computer interaction (HCI) systems used in real life applications. The target is to create machines and robots that are capable of understanding people's feelings, emotions and behaviors; thus, being able to interact in a 'human-centered', and effectively serving them as their digital assistants. Human affective behavior analysis plays a significant role in HCI systems. 

Different action units (AU) combinations in Facial Action Coding System (FACS)\cite{ekman1978facial} can represent different expressions. The FACS defines a set of facial action units from the perspective of face anatomy, which is used to accurately characterize the facial expression changes. Each facial action unit describes a set of apparent changes generated by facial muscle movements, the combination of which can express arbitrary face expressions. Facial action units (AUs) relate to specific local facial regions based on the FACS, so how to effectively extract the local features associated with the corresponding AU is particularly important. Traditional methods \cite{chu2013selective,ding2013facial,eleftheriadis2015multi,wang2013capturing,liu2013aware} use handcrafted methods to represent facial local regions. With the development of deep learning, deep neural networks have been used to improve the accuracy of AU detection, and use face landmarks or divide aligned faces into different patches to locate facial local areas. Obviously, the above two methods fixedly extract the local facial region, which is not accurate enough and cannot adapt to the posture changes of the face. Recent work \cite{tang2021piap} has emerged that uses a three-stage training strategy to adaptively enable the encoder to extract features that perceive facial local regions. However, this method still needs to use the extra annotations related to the face landmarks, and use multi-task learning to train the model. Because the activation status of AUs are not independent of each other, the activation status of one AU is often correlated with the status of other AUs. Therefore, the relationship between AUs should be taken into consideration when performing AU detection. A recent work\cite{luo2022learning} uses a graph neural network to obtain the relationship between AUs, and through a two-stage training strategy, obtains multi-dimensional edge features. However, in order to obtain the relationship between AU nodes, using a two-stage training strategy makes the training process complicated, therefore, a more effective module is used in our proposed method to extract the features of AU nodes.

In this paper, We propose an end-to-end method for AU detection. It can more effectively and adaptively consider the facial local regions related to AU detection, so that it only needs to be trained once to extract the desired relationship between AU nodes and output the relation graph of AU. Considering that the overall characteristics of the target face ( such as whether the expression is happy or sad. ) also plays a certain role in AU detection, the influence of the overall representation of the face will eventually be integrated. Specifically, our method consists of three modules: (i) the \textbf{Local region perception (LRP) module} effectively extracts the parts relevant to AU detection in the output of backbone; and the (ii) \textbf{AU relationship learning (ARL) module} learns the relation representation between AUs; the (iii) \textbf{feature fusion (FF) module} fuses the mutual information between AUs and the overall representation of the target face.

In this work, we focus on Action Unit Detection.
Our contributions in this paper are summarized as: 
\begin{itemize}
    \item We use the LRP module to make the model better capture the facial local region related to AU detection without using additional facial 
    landmark annotation and multi-stage training process.
\end{itemize}
\begin{itemize}
    \item The ARL module based on the graph neural network is used to learn multiple relationship graphs between AUs.
\end{itemize}
\begin{itemize}
    \item In order to better fuse the overall features of the target face and the relationship features between the AUs, We propose a feature fusion module based on self-attention operations, which achieves the best result on the official validation set, but on some other cross-validation sets, simple fixed weight fusion can achieve better result.
\end{itemize}

\begin{figure*}[ht]
\centering
\includegraphics[width=\linewidth]{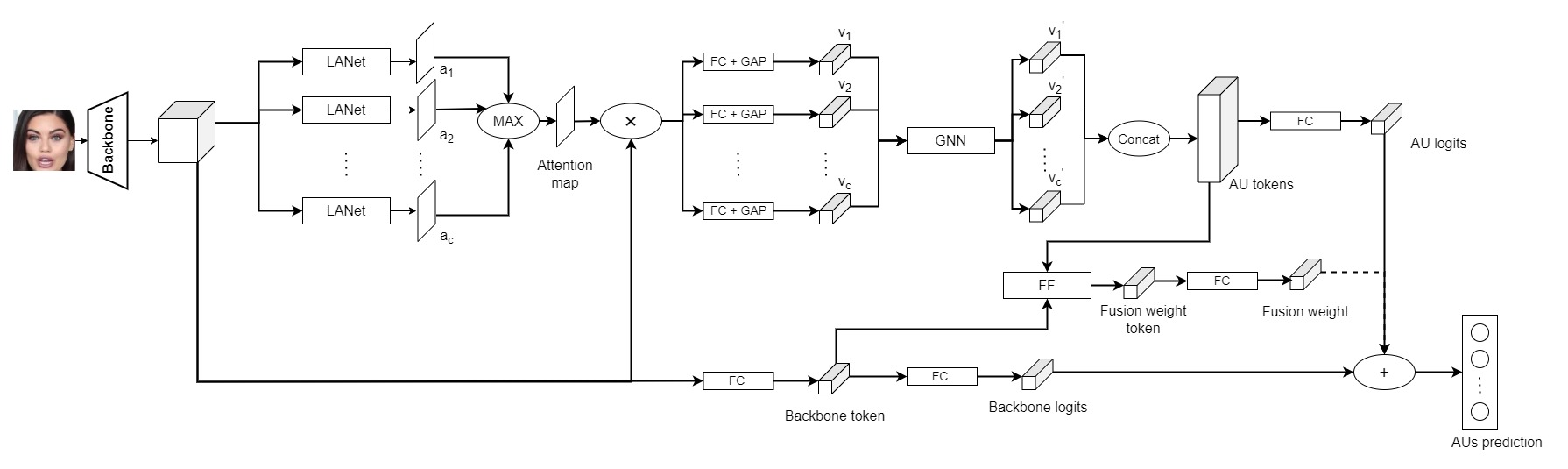}
\caption{Our proposed framework for AU detection. The backbone uses the input target face to extract the overall features of the image. The local region perception module helps the graph neural network to more effectively extract the relationship between AU nodes, the relationship learning module learns the correlation between different AUs, and the feature fusion module considers the overall information of the target face thereby helping AU detection. In addition, we only draw the feature fusion module based on self-attention, and do not draw the simple fixed weight fusion.}
\label{Figure 1}
\end{figure*}

\section{Related works}
\label{sec:formatting}

In this section, we shortly summarize some works related to the problem of AU detection in the challenge.

\subsection{Competition on ABAW}
The competition on affective behavior analysis in-the-wild is dedicated to solving the problem of computer analysis of human emotion behavior in natural situations, and thus improving the scenario application capability of human-computer interaction systems. The competition was held at this year's International Conference on Computer Vision and Pattern Recognition (CVPR).

In the previous challenge\cite{kollias2022abaw}, many effective approaches offered by some scholars. For example, Wei Zhang et al.\cite{zhang2022transformer} propose a unified transformer-based multimodal framework for Action Unit detection. Lingfeng Wang et al.\cite{wang2022multi} propose a transfomer\cite{vaswani2017attention} based model to detect facial action unit (FAU) in video. They propose a action units correlation module to learn relationships between each action unit labels and refine action unit detection result.We noted that applying a multi-label detection transformer\cite{vaswani2017attention} that leverage multi-head attention to learn which part of the face image is the most relevant to predict each AU, which is a very effective solution. These methods have given a great boost to the development of AU detection tasks.

\subsection{AU detection}
For the task of AU detection, the limited identity of commonly used datasets and the inability to extract local features relevant to each AU detection are major challenges. 
Due to the complexity of AU labeling, traditional methods represent local areas of the face by handcrafted has significant limitations.
In order to solve the above problems, many recent works focus on using additional facial landmarks annotations to extract important facial local features, and use multi-task learning to improve the performance of AU detection models \cite{benitez2017recognition, shao2021jaa, jacob2021facial}.
In SEV-Net\cite{yang2021exploiting}, text descriptions of local information are used to generate local region attention maps. 

In order to enable the model to adaptively learn features that carry key facial local region information, Tang \emph{et al.} \cite{tang2021piap} adopts a three-stage training strategy, using face landmarks information in a way similar to multi-task learning, so that the model can pay attention to those important facial local regions. However, this method requires the use of additional face landmarks annotations, and does not pay attention to the possible associations between AUs. Luo \emph{et al.} \cite{luo2022learning} adopted a two-stage training strategy based on a graph neural network to obtain the associated state relationship between AUs. However, this method only uses a simple fully connected layer to obtain the characteristics of each AU node and doesn't use additional face key landmarks annotation as an auxiliary, so an additional first stage of training is required to enable the fully connected layer to obtain node information well. 
In short, these methods try to introduce additional tasks or annotations to help the model learn important facial local region features.
Thus, The proposed method uses a local region perception (LRP) module to optimize the global features output from the backbone network, making it easier for the fully connected layers to learn the features of each AU node. And we use a more simplified graph neural network than in \cite{luo2022learning} and a strategy of fusing with the global features output by the backbone network. Next, we will describe our work more specifically.

\begin{figure}[ht] 	
\centering
\includegraphics[scale=0.5]{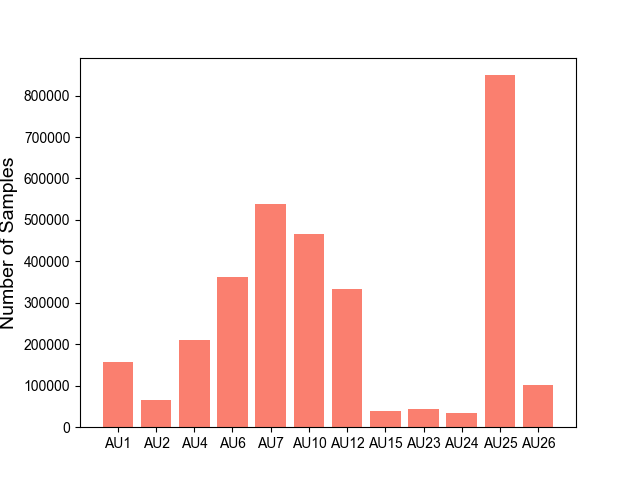} 	
\caption{The distribution of the number of AUs in each category in the training set.}
\label{Figure 2}		
\end{figure}

\section{Method}
The architecture of the proposed AU detection framework is shown in Fig. \ref{Figure 1}. The entire framework includes a feature extractor (IResnet100\cite{duta2021improved}) pre-trained on Glint360K\cite{an2021partial}, a local region perception module, a AU relationship learning module and a feature fusion module.
The feature extractor extracts the overall representation of the input image, and then the output of the backbone is input to two branches. One branch effectively extracts the part feature related to AU detection in the output of the backbone through the local attention module, and the optimized output is passed through AU relationship learning module based on graph neural network acquires relational representations between AUs.
The other branch is input in parallel to a fully connected layer and a feature fusion module, and the fully connected layer is used to obtain the logits output by the backbone network. We finally chose two types of feature fusion modules, which achieved the best results on different cross-validation sets. One is based on self-attention operation, considering the relationship between the overall features of the target face and the AUs feature graph. This type of feature fusion module obtained the best result on the official verification set; the other is to simply fuse the logits output by the backbone with the logits output by the graph neural network with a fixed weight, and this type of feature fusion module has achieved the best results in some cross-validation sets.
\subsection{Local region perception module}
In order to help the graph neural network extract mutual information between AU nodes effectively, we propose the Local region perception module. This module consists of several LANets\cite{wang2019ls}, which effectively notice the local region of the face associated with AU detection. LANet consists of two 1x1 convolutional layers. Assuming that the feature dimension output by the backbone is (c, h, w), after the first 1x1 convolution, the number of channels will be reduced to c/r, where r represents the channel compression rate; after the second 1x1 convolution, the number of channels will be reduced to 1. Several feature maps output by LANets will be stacked together in the channel dimension, and also the maximum value will be calculated in the channel dimension, and finally the attention score map will be obtained through the sigmoid activation function.
The attention score map finally output by the local region perception module and the output of the backbone are element-wise multiplied to obtain the required feature map related to AUs.
\subsection{AU relationship learning module}
The AU relationship learning module uses the FGG network proposed in \cite{luo2022learning} to obtain the association information between AUs. The optimized backbone network output obtained from the previous module will first go through the number of categories of fully connected layers and global average pooling to obtain the features $\bm{\mathcal{V}} = \{v_{1}, v_{2}, ..., v_{c}\}$ of each AU node. Then the AU node features are input into the graph neural network to obtain the relationship information between AUs. The graph neural network generates a specific topology for each target face, and allows multiple relationships between nodes, and the number of relationships is determined by the hyperparameter k. The graph neural network will output the features $\bm{\mathcal{V}^{'}} = \{v_{1}^{'}, v_{2}^{'}, ..., v_{c}^{'}\}$ of each AU node, which already contains the mutual information between the node and k other AU nodes, and finally stack the features of all AU nodes together through the fully connected layer to get AU logits.

\begin{figure}[ht]
\centering
\includegraphics[scale=0.5]{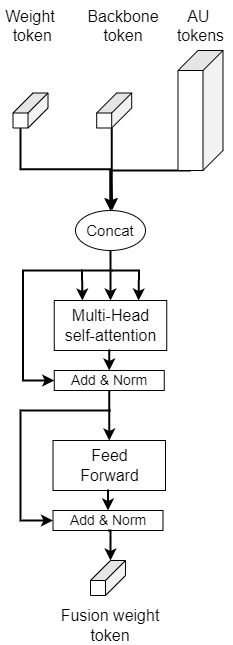}
\caption{Feature fusion module based on self-attention operation. The weight token is similar to the class token in Vision Transformer\cite{dosovitskiy2020image}, which is a learnable vector. The weight token, backbone token and AU tokens concat together as q, k, v are input into the encoder layer to obtain the fusion weight token used to output the fusion weight. In the experiment, the number of encoder layer is 2.}
\label{Figure 3}
\end{figure}

\begin{table*}[]
\resizebox{\textwidth}{!}{
\begin{tabular}{@{}l|llllllllllll|l@{}}
\toprule
\textbf{Val Set} & \textbf{AU1}   & \textbf{AU2}   & \textbf{AU4}   & \textbf{AU6}   & \textbf{AU7}   & \textbf{AU10}  & \textbf{AU12}  & \textbf{AU15}  & \textbf{AU23}  & \textbf{AU24}  & \textbf{AU25}  & \textbf{AU26}  & \textbf{Avg.}                \\ \midrule
Official         & 56.41          & 50.94          & 54.46          & 60.03          & 71.00          & 71.61          & 70.30          & 30.41          & \textbf{14.19} & 15.26          & \textbf{80.56} & 37.52          & 51.06                        \\
Fold-1           & 57.95          & \textbf{53.48} & \textbf{47.45} & 60.67          & \textbf{68.42} & \textbf{72.07} & 70.28          & \textbf{39.94} & \textbf{26.66} & 13.37          & 80.63          & \textbf{45.28} & {\textcolor{red}{53.02}} \\
Fold-2           & \textbf{48.50} & 31.26          & 62.38          & \textbf{65.15} & \textbf{71.97} & 70.71          & \textbf{75.39} & 38.85          & 22.72          & \textbf{22.50} & \textbf{82.34} & 30.90          & 51.97                        \\
Fold-3           & 48.84          & \textbf{26.50} & \textbf{63.52} & 60.63          & 69.31          & \textbf{69.02} & 70.03          & 32.04          & 21.34          & 19.41          & 81.24          & \textbf{26.76} & 49.05                        \\
Fold-4           & \textbf{59.04} & 41.75          & 49.89          & \textbf{58.23} & 69.90          & 69.71          & \textbf{66.63} & \textbf{17.43} & 25.04          & \textbf{9.69}  & 81.70          & 40.74          & \textcolor{red}{49.15} \\ \bottomrule
\end{tabular}}
\caption{The AU F1 scores of models that are trained and tested on different folds (including the official training/validation set). The highest and lowest scores are both indicated in bold.}
\label{val result table}
\end{table*}

\subsection{Feature fusion module}
In order to consider the overall information of the target face at the same time, the proposed feature fusion module can add the Backbone logits and the AU logits according to a certain fusion weight. In the competition, we used two methods to obtain fusion weights. One method simply uses fixed weights to add AU logits and Backbone logits. Another method based on the self-attention operation is shown in Fig. \ref{Figure 3}. The AU tokens and Backbone token obtained by passing the output of the backbone network through a fully connected layer are stacked together. Similar to the approach in Vision Transformer \cite{dosovitskiy2020image}, we add at the beginning of the sequence a Weight token, which can get the fusion weight through the fully connected layer. Finally, the fused logits pass through the sigmoid activation function to obtain the predicted probabilities of each category.

\subsection{Training}
Some traning details and tricks are introduced in this part.

\subsubsection{Resampling}
Since the dataset is composed of continuous video frames, the category changes and image changes of adjacent video frames are small. Therefore, in order to reduce the training time and avoid the model from quickly overfitting to the training set in a few epochs, we uniformly sample one-tenth of the pictures as the training set. In addition, due to the serious category imbalance in the dataset shown in Fig. \ref{Figure 2}, we use a uniform sampling strategy of one-fifth of the five categories of AU2, AU15, AU23, AU24, and AU26.
\subsubsection{Loss function}
AU detection is a multi-label classification task. We use two loss functions, namely binary cross-entropy (bce) loss and circle loss for multi-label classification. Its calculation formula is as follows:
\begin{equation}
    \mathcal{L}_{bce} = -\frac{1}{12}\sum_{j=1}^{12}[\;y_{j} log\hat y_{j} + (1 - y_{j}) log(1 - \hat y_{j})\;]
\end{equation}
\begin{equation}
    \mathcal{L}_{circle} = log(1 + \sum_{i\in\Omega_{neg}}e^{s_{i}}) + log(1 + \sum_{j\in\Omega_{pos}}e^{-s_{j}})
\end{equation}
\begin{center}
    $\Omega_{neg} = \{ \; i\;|\;if\;y_{i} = 0 \}$
\end{center}
\begin{center}
    $\Omega_{pos} = \{ \; j\;|\;if\;y_{j} = 1 \}$
\end{center}
Finally, add bce loss and circle loss together, as (3).
\begin{equation}
    \mathcal{L}_{total} =  \mathcal{L}_{bce} + \mathcal{L}_{circle}
\end{equation}

\begin{table}[]
\resizebox{\linewidth}{!}{
\begin{tabular}{@{}lllllll@{}}
\toprule
IResNet100 & \begin{tabular}[c]{@{}l@{}}Circle \\ loss\end{tabular} & \begin{tabular}[c]{@{}l@{}}Glint360K\\ pre-train\end{tabular} & MIL & LRP & FF & \begin{tabular}[c]{@{}l@{}}F1 score\\ (Official)\end{tabular} \\ \midrule
\checkmark          &                                                        &                                                               &     &     &    & 46.82                                                         \\
\checkmark          & \checkmark                                                     &                                                               &     &     &    & 47.78                                                         \\
\checkmark          & \checkmark                                                      & \checkmark                                                             &     &     &    & 49.06                                                         \\
\checkmark          & \checkmark                                                      & \checkmark                                                             & \checkmark   &     &    & 49.56                                                         \\
\checkmark          & \checkmark                                                      & \checkmark                                                             & \checkmark   & \checkmark   &    & 50.19                                                         \\
\checkmark          & \checkmark                                                      & \checkmark                                                             & \checkmark   & \checkmark   & \checkmark  & 51.06                                                         \\ \bottomrule
\end{tabular}}
\caption{Ablation experimental results of our proposed method and modules on the official validation set.}
\label{ablation table}
\end{table}

\section{Experiments}
In this part, we will first introduce the dataset used in this competition. We then present our implementation details and result on the validation set. Finally, to demonstrate the effectiveness of the aforementioned modules, we present the results of ablation experiments.
\subsection{Datasets}
For this Challenge, the Aff-Wild2 dataset will be used.
The Aff-wild2\cite{kollias2018aff,kollias2022abaw,kollias2021distribution,kollias2019expression,kollias2021affect,kollias2021analysing} was extended from Aff-wild1\cite{zafeiriou2017aff, kollias2020analysing, kollias2019deep, kollias2019face}. Aff-wild2 expand the number of videos with 567 videos annotated by valence-arousal, 548 videos annotated by 8 expression categories, 547 videos annotated by 12 AUs, and 172,360 images are used that contain annotations of valence-arousal; 6 basic expressions, plus the neutral state, plus the ’other’ category; 12 action units. The Action Unit Detection task includes 548 videos annotating the 6 basic expressions, plus the neutral state, plus a category ’other’ that denotes expressions/affective states other than the six basic ones. Approximately 2,6 million frames, with 431 participants (265 males and 166 females), have been annotated by seven experts. Therefore, the Aff-Wild2\cite{kollias2018aff,kollias2022abaw,kollias2021distribution,kollias2019expression,kollias2021affect,kollias2021analysing} show human spontaneous affective behaviors in the wild, pushing the affective analysis to fit with the real-world scenarios.
\subsection{Implementation details}
We used IResnet100 pre-trained on Glint360k as the backbone, and the other modules of the model were trained from scratch. The whole training process consists of 15 epochs, the initial learning rate is 0.001, the stochastic gradient descent algorithm is used and the batch size is 256. The learning rate is reduced to one-tenth of the original at the 4th, 6th, and 8th steps. For data enhancement, we only use commonly used weak data enhancements, such as horizontal flipping and color jitter, instead of strong data enhancements such as MixUp, because it will conflict with the loss function we use.
\subsection{Results on validation set}
In order to make full use of the official datasets provided. In addition to the official validation set, we also did 4-fold cross-validation. The results of these five partitioned validation sets are shown in Table \ref{val result table}. Some of the best results on the validation set were obtained with a fixed-weight feature fusion module, which we denote in red. Finally, we use the five models to vote on the test set as the final prediction.

\subsection{Ablation study}
In order to prove the effectiveness of the proposed method and module, we have done detailed ablation experiments shown in table \ref{ablation table}, and the results of all experiments are carried out on the official validation set. Our baseline is obtained by adding a classification head after IResNet100. When using Circle loss, the F1 score increased by about 1\%; when IResNet100 was loaded with pre-trained weights on glint360k dataset, the F1 score increased by 1.5\%; after using the ARL module, the model learned the relationship between AU, the F1 score has increased by about 0.5\%; after adding the LRP module, the model can better extract key local area features, so the score has increased by 0.63\%; after finally integrating the overall features of the target face output by the backbone network, the F1 score has increased by 0.83\%.

\section{Conclusion}
In this paper, we introduce the proposed single-stage AU detection framework for the AU detection challenge in the ABAW5 competition.
Aiming at the extraction of local region features of AU detection tasks and the relationship learning problem between AUs, we adopted a local region perception module based on LANet and a relationship learning module based on graph neural network.
In addition, in order to integrate the overall features of the target face and the relationship features between the AUs, we employ a feature fusion module based on self-attention operations.
The results of ablation experiments show that the modules we use can improve the performance of AU detection model.

\section{Acknowledgments}
This work was supported by the Natural Science Foundation of China (62276242), CAAI-Huawei MindSpore Open Fund (CAAIXSJLJJ-2021-016B, CAAIXSJLJJ-2022-001A), Anhui Province Key Research and Development Program (202104a05020007), USTC-IAT Application Sci. \& Tech, Achievement Cultivation Program (JL06521001Y), Sci.\&Tech and Innovation Special Zone (20-163-14-LZ-001-004-01).

{\small
\bibliographystyle{ieee_fullname}
\bibliography{egbib}
}

\end{document}